# A Lightweight and Explainable DenseNet-121 Framework for Grape Leaf Disease Classification

Md. Ehsanul Haque
Dept. of CSE
East West University
Dhaka, Bangladesh
2021-3-60-018@std.ewubd.edu

Md.Saymon Hosen Polash
Dept. of CSE
East West University
Dhaka, Bangladesh
2022-1-60-171@std.ewubd.edu

Rakib Hasan Ovi
Dept. of CSE
East West University
Dhaka, Bangladesh
2022-1-60-342@std.ewubd.edu

Aminul Kader Bulbul*
Dept. of CSE
East West University
Dhaka, Bangladesh
aminul.bulbul@ewubd.edu

Md Kamrul Siam
Dept. of Computer Science
New York Institute of Technology
New York, USA
ksiam01@nyit.edu

Tamim Hasan Saykat
Dept. of CSE
East West University
Dhaka, Bangladesh
2022-1-60-289@std.ewubd.edu

*Abstract*—Grapes are among the most economically and culturally significant fruits on a global scale, and table grapes and wine are produced in significant quantities in Europe and Asia. The production and quality of grapes are significantly impacted by grape diseases such as Bacterial Rot, Downy Mildew, and Powdery Mildew. Consequently, the sustainable management of a vineyard necessitates the early and precise identification of these diseases. Current automated methods, particularly those that are based on the YOLO framework, are often computationally costly and lack interpretability that makes them unsuitable for real-world scenarios. This study proposes grape leaf disease classification using Optimized DenseNet 121. Domain-specific preprocessing and extensive connectivity reveal disease-relevant characteristics, including veins, edges, and lesions. An extensive comparison with baseline CNN models, including ResNet18, VGG16, AlexNet, and SqueezeNet, demonstrates that the proposed model exhibits superior performance. It achieves an accuracy of 99.27%, an F1 score of 99.28%, a specificity of 99.71%, and a Kappa of 98.86%, with an inference time of 9 seconds. The cross-validation findings show a mean accuracy of 99.12%, indicating strength and generalizability across all classes. We also employ Grad-CAM to highlight disease-related regions to guarantee the model is highlighting physiologically relevant aspects and increase transparency and confidence. Model optimization reduces processing requirements for real-time deployment, while transfer learning ensures consistency on smaller and unbalanced samples. An effective architecture, domain-specific preprocessing, and interpretable outputs make the proposed framework scalable, precise, and computationally inexpensive for detecting grape leaf diseases.

## I. INTRODUCTION

The global viticulture industry plays a crucial role in both agricultural sustainability and economic development. The grape serves as the base for table fruit and global wine production, including Europe and Asia. [1]. Based on recent statistics, Europe is a principal wine producer, with Italy, France, and Spain as the foremost contributors globally [2]. Additionally, Asia is witnessing a surge in wine and table grape production, with countries such as China, India, and Turkey gaining prominence [3], [4]. Even though grape yield and quality is so important, the grape is very sensitive to leaf diseases like Bacterial Rot, Downy Mildew and Powdery Mildew [5], [6]. The timely and precise diagnosis of these diseases is essential to the reduction of yield losses, the decrease of pesticides application, and the encouragement of the sustainable management of vineyards [7], [8]. Though traditional, visual inspection is a labor intensive system that is time consuming and can result in human error especially in extensive agricultural operations [9], [10]. The advancement of computer vision and deep learning has transformed the methodology for detecting plant diseases and facilitating growth by offering automated and scalable solutions [11], [12]. Convolutional Neural Networks (CNNs) and vision transformers have proven to be very accurate in classifying diseases in plants within different crops. By utilizing knowledge from pretrained models, feature extraction becomes more robust, facilitating accurate disease classification in small grape leaf datasets [13]. However, major issues such as small and unbalanced datasets, leaf orientation and lighting variability, and big computational cost of deep models can still limit real-time implementation. Moreover, the lack of transparency in most deep learning models hampers their interpretability and undermines trust, restricting their practical application in viticulture. Furthermore, many studies use object detection models such as YOLO. But YOLO-based object detection frameworks are popular, they often incur high computational costs and slower inference, highlighting the need for models that balance efficiency, accuracy, and interpretability [14], [15]. To overcome these constraints, this research proposes an Optimized DenseNet 121 network in grape leaf disease detection. The framework uses domain specific preprocessing, such as resizing, enhancement of green channel, contrast adjustment, noise removal and normalization to highlight features of disease importance. Compared to existing models, the dense connectivity facilitates the network to effectively extract low level features, including veins and edges, high level lesion patterns, and optimization can minimize the cost of computation, making the model feasible in real time monitoring. Transfer learning also boosts the generalization of small training data, given that transfer learning guarantees realistic usefulness in a variety of conditions in vineyards. Grad CAM based explainable AI is employed to highlight regions that contribute most to the model predictions. These visualizations verify that the model is focus on biologically meaningful area, i.e., lesions and discolorations, which enhances transparency and builds confidence in automated decision making. It is thoroughly tested and compared to a variety of baseline CNN architectures, demonstrating superior performance and computational efficiency. Through a combination of powerful preprocessing, efficient architecture, and interpretable results, this paper provides a scalable, precise, and computationally efficient grape leaf disease detector, which meets the key gaps in research and practice of viticulture in Europe, Asia, and other countries.

## II. RELATED WORKS

Recent developments in computer vision and deep learning have greatly boosted studies on automated grape leaf disease detection. In this section, we present recent literature on grape leaf disease detection, highlighting both their contributions and limitations. Prasad et al. [16] extended upon this work with a VGG16 based DCNN and reported 99.18% training accuracy and 99.06% testing accuracy classification for grape leaves. Despite the strengths of the work, situating the evaluation of performance against input datasets that exist in silico does not adequately account for variability found in field variability. Meanwhile, Zhang et al. [17] developed DLVTNet, a lightweight model that combines Ghost convolutions with Transformer modules. DLVTNet was assessed within the framework of the New Plant Disease dataset on which it achieved an accuracy score of 98.48%. While model performance within these confines has potential value, performance against real world use is not verified.Expanding upon transformer based methods, Lu et al. [18] developed GeT, which was based on a Ghost Transformer model that achieved 98.14% accuracy with a speed of 180 FPS with only 1.16M of parameters. However, it only tested using the GLDP12k dataset and the generalisation to new environments is uncertain. In grape bunch detection, Pinheiro et al. [19] used YOLO to achieved 77% mAP for bunch detection and 92% F1 score for condition classification. While this is promising, the mAP is still less than 80%, which shows there is still room for improvement. Similarly, Yang et al. [20] proposed YOLOv8s-grape which was a lightweight version of YOLOv8s, with +2.4% mAP50–95 improvement over the previous YOLOv8s, while improving computational cost. However, the evaluation was applied to constructed datasets only, which needs to be tested in actual situations to determine robustness in diverse conditions. Zhang et al. [21] introduced YOLOv5-CA for grape downy mildew detection, and achieved an mAP of 89.55% at a speed of 58.8 FPS. While it was able to outperform all other tested models on the vineyard dataset, their narrow focus on a single disease type and a single recall rate, shows their model still needs to improve further. Atesoglu and Bingol [22] used a combination of CNN and texture features (from DenseNet201) with Neighborhood Component Analysis (NCA), which achieved a 99.1% accuracy. Although it showed promising accuracy, this method was vastly complex, and no explicit evaluations on field performance were reported. Kunduracioglu and Pacal [23] also conducted a more recent study on similar elements in grape vine health.Nonetheless, the small dataset raises concerns about the generalizability of these findings. Karim et al. [24] adapted MobileNetV3Large for on-device grape disease classification application and achieved a 99.42% test accuracy. The model does show feasibility at the edge; however, it has not been tested for real-world accuracies, nor has it been characterized for inference latency. Shetty [25] also compared convolutional neural networks (CNN) and DenseNet for grape leaf classification, showing a 97% accuracy for DenseNet. The authors did not provide extensive detail to support generalizability, nor did they perform external validation of their study. In another comparison study, Om et al. [26] assessed performance of seven pre-trained CNNs, reporting by DenseNet121 as outperforming on the PlantVillage dataset with 99.21% accuracy. They did not evaluate performance in real-world conditions, thus remaining unclear as to the robustness of their findings. Finally, Patil and More [27] compared five deep learning models for grape leaf diseases, achieving 99.86% accuracy and recall on DenseNet121. Unfortunately, while the performance was impressive, the dataset was small, and no real world testing occurred, thus limiting application in practice.

### A. Identified Gaps in Current Approaches

YOLO-based models are very effective, yet they can be heavy and slow, and therefore often impractical for real-time applications. There are many existing methods whose computational expense hinders the use of available mobile devices. Lastly, due to the lack of Explainable AI (XAI) integration, there is no transparency or trust in prediction made in the agricultural area by these models.Additionally, the question of overfitting controlled datasets remains. These models struggle with generalization and the variability of the real-world including lighting, leaf orientation, and environmental elements.

## III. METHODOLOGY

This research proposes a systematic approach to the problem of computational inefficiency, poor generalization, and lack of interpretability proposed in the current literature by incorporating dataset preparation, domain specific preprocessing, model design and evaluation, as shown in Figure 1. The entire workflow contributes to accuracy as well as real-world applicability of grape leaf disease detection.

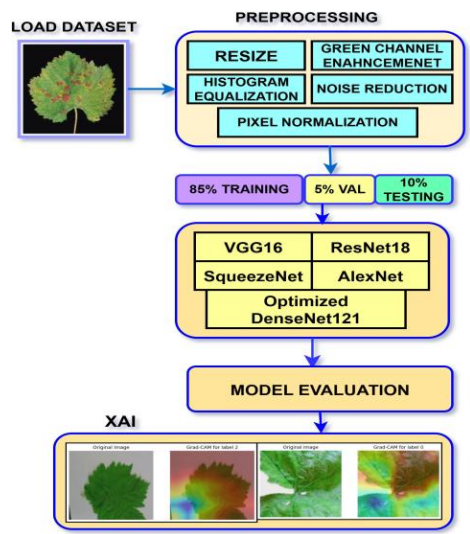

Fig. 1. Workflow for the proposed grape leaf disease detection framework

### A. Data Collection

The dataset used for grape leaf disease detection was collected from the Mendeley Niphad Grape Leaf Disease Dataset, which contains images of grape leaves showing different types of diseases as well as healthy leaves [28], [29]. The dataset is imbalanced and some classes have significantly more samples than others, which poses a challenge for model training. Figure 2 presents the class-wise distribution of images used in this study.

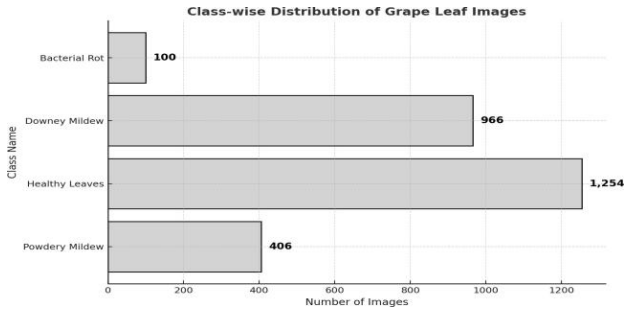

Fig. 2. Class-wise distribution of grape leaf images in the dataset where Healthy leaves has the higher image count whereas bacterial rot has the lowest.

Moreover, Figure 3 shows some sample image from the NIPHAD Dataset.

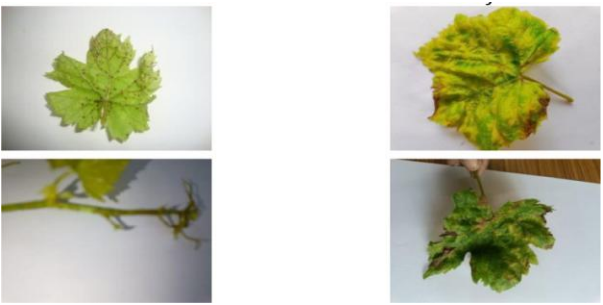

Fig. 3. Sample images from the NIPHAD dataset showing grape leaves affected by different diseases and healthy leaves.

### B. Preprocessing

Before training, the grape leaf images were preprocessed to standardize the dataset and increase the reliability of disease related features. All images were resized to uniformly be 224 × 224 pixels prior to training to create efficiency and provide consistent size of as well for batch processing training. The green channel of the leaf was enhanced using contrast stretching to highlight various disease related features, while histogram normalization was also conducted to add more contrast to distinguish lesions and spots on the leaf. Noise was reduced by using median filters, while preserving important leaf structures. Finally, pixel values were normalized into the range [0,1], and normalization was also done on a per channel basis using the appropriate ImageNet statistics used to speed up training. Class labels, initially in categorical form (e.g., *Bacterial Rot*, *Healthy Leaves*), were converted to numerical indices to ensure compatibility with classification algorithms. Each label was mapped to an integer, which was applied consistently across the training, validation, and test datasets, enabling efficient model learning and evaluation.

### C. Dataset Splitting

The dataset demonstrates the need for sound evaluation practice and keeping track of class distributions across splits. We performed the splits using a stratified splitting strategy, ensuring the distribution of the data was comparable. An exhaustive form of perfectly reproducible stratification was done on the entire dataset. Data was initially split 80 % for training, leaving 20% aside for the time being. This 20% temporarily set aside dataset, included a validation subset (5% of the total) and a test subset (15% of the total); and also ensured that the class distributions remained the original proportions.

The class-wise distribution of images in each split is summarized in Table I.

TABLE I
CLASS-WISE DISTRIBUTION OF GRAPE LEAF IMAGES IN TRAINING, VALIDATION, AND TESTING SETS

| Class Name | Training | Validation | Testing | Total |
| --- | --- | --- | --- | --- |
| Healthy Leaves | 1003 | 63 | 188 | 1254 |
| Downey Mildew | 772 | 48 | 146 | 966 |
| Powdery Mildew | 325 | 20 | 61 | 406 |
| Bacterial Rot | 80 | 5 | 15 | 100 |

### D. Baseline Classifiers

In order to benchmark the proposed grape leaf disease classification framework, a number of established deep learning models were chosen and trained on the dataset via transfer learning. Beginning with DenseNet121, its ability to efficiently reuse features and mitigate vanishing gradients enables stronger performance, particularly on smaller datasets. Building on the concept of efficiency, SqueezeNet was included for its minimal number of parameters, allowing rapid training and inference while maintaining high accuracy. In contrast, AlexNet provides a simpler yet historically influential architecture, serving as a baseline for comparison. Moving to deeper architectures, VGG16 captures hierarchical features through its extensive layers, whereas ResNet18 leverages skip connections to balance depth, accuracy, and computational cost. Comparing these models provides a detailed understanding of their relative strengths and limitations, thereby highlighting the advantages of the proposed framework in achieving accurate and efficient grape leaf disease classification.

### E. Proposed Optimized DenseNet-121

We propose an optimized DenseNet-121 based framework to classify grape leaf diseases. DenseNet-121 effectively reuses features and retains strong gradient flows which makes it a good fit for moderate size datasets. The model is designed with four dense blocks with transition layers that learn and extract both low-level features as well as more complex high level disease features.

The model is trained using pretrained ImageNet weights and fine-tuned with a fully connected classification head corresponding to the number of target classes. To prevent overfitting, dropout layers and L2 regularization were incorporated. The final layer employs a softmax activation function to output probabilities for the four categories: Bacterial Rot, Downy Mildew, Powdery Mildew, and Healthy Leaves. Training is performed using cross-entropy loss with the Adam optimizer, and a learning rate scheduler guided by validation metrics ensures optimal performance during inference. This framework effectively captures subtle disease patterns, leading to high classification accuracy. Figure 4 illustrates the main architectural components and their roles in extracting hierarchical features for grape leaf disease classification.

### F. Model Training Settings

Key training settings for all evaluated models for classifying grape leaf diseases are summarized in Table II.

### G. Explainable AI

Grad CAM was applied to the proposed DenseNet-121 model in order to ensure that the most relevant regions of the grape leaf images were effectively visualized. The heatmaps generated by Grad-CAM highlighted key disease features such as lesions, discoloration, and powdery spots. That confirms the model focused on meaningful areas. By overlaying the heatmaps on the original images, we could validate that the model's predictions were based on relevant disease symptoms.

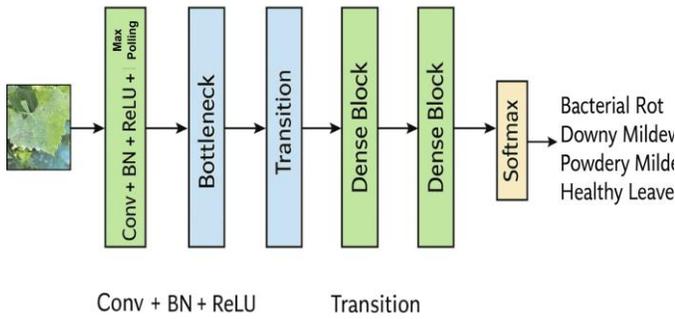

Fig. 4. Overview of the proposed optimized DenseNet-121 framework for grape leaf disease classification.

TABLE II
TRAINING SETTINGS FOR DENSENET-121

| Parameter | Setting |
|---|---|
| Optimizer | Adam |
| Loss Function | Cross-Entropy Loss |
| Learning Rate | 0.001 |
| Learning Rate Scheduler | Step/Adaptive (based on validation loss) |
| Epochs | 20 (maximum) |
| Early Stopping | Patience = 5, Minimum Delta = 0.001 |
| Batch Size | 32 |
| Mixed Precision | Enabled (AMP with GradScaler) |
| Regularization | Dropout = 0.4 + L2 Weight Decay |

## IV. RESULTS AND DISCUSSION

In this section we are now going to compare the performance of all the models that have been evaluated. The evaluation metrics included are accuracy, precision, recall, F1 score, MCC, PR AUC, Kappa, specificity, and training and inference time of each model. These evaluation metrics are used to analyze the effectiveness of the models in classifying the grape leaf diseases and the overall effectiveness of the models. The findings give a full picture of the strengths and weaknesses of both models.

### A. Evaluation of Model Performance

Table III summarizes the performance of all evaluated models for classifying grape leaf diseases. The proposed Optimized DenseNet121 was best out of all evaluated models, in terms of its performance in all metrics. In particular, it achieved an accuracy of 99.27%, recall of 99.27%, precision of 99.29% and F1 of 99.28%. It also had the best performance regarding MCC (98.86%), PR AUC (99.99%), Kappa (98.86%), and specificity (99.71%). These findings illustrate the high capability of DenseNet121 to successfully predict grape leaf diseases correctly classifying them into different classes of diseases. Comparatively, ResNet18 also achieved a good score with an accuracy of 97.56%, recall of 97.81%, precision of 97.81%, and specificity of 99.21%. Although its performance was good, it was marginally lower than optimized DenseNet121 on many key metrics, such as the F1 score and the PR AUC. The other models, VGG16, AlexNet and SqueezeNet, performed poorly compared to DenseNet121 and ResNet 18. The accuracy of VGG16 was 96.34% whereas AlexNet and SqueezeNet had an accuracy of 94.39% and 95.12%, respectively. Despite the similar precision and recall of

These models were significantly lower than the highest-performing models. To sum up, the proposed Optimized DenseNet-121 model had the best overall results as it was able to achieve a better classification accuracy and strength in classification of grape leaf diseases. We also cross validated the proposed model by using 5 folds and the figure below depicts the cross-validation accuracy of the proposed model across the 5 folds.

Figure 5 below presents the cross-validation accuracy of 5 folds of the proposed Optimized DenseNet121 model. The model was consistent and showed steady performance with the accuracy of between 98.84% and 99.05%. The fourth fold had the highest accuracy of 99.05% and the second fold had the lowest accuracy of 98.84%. These findings indicate the strength and generalizability of the proposed Optimized DenseNet-121 model which makes it reliable when considering various data splits.

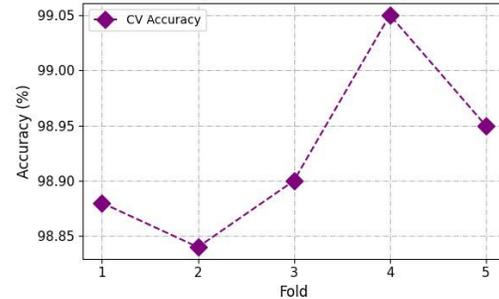

Fig. 5. Cross-validation accuracy across 5 folds for the proposed Optimized DenseNet-121 model.

### B. Optimized DenseNet121 Convergence Analysis (Loss and Accuracy Curves)

Figure 6 shows that the optimized DenseNet121 converges very quickly during the initial few epochs as both training and validation losses decrease drastically. After this stage, the curves become flat, which means that the model has entered a steady optimization regime. It is remarkable that validation accuracy follows the training accuracy closely, and it can be concluded that the model effectively generalizes on unseen data rather than just memorizes the training set. The fact that accuracy reaches nearly 99% almost at its peak, supports the architecture efficiency of feature discrimination. Moreover, the lack of a significant difference in the training and validation curves means that there is a minimal overfitting.

### C. Misclassification Analysis of Optimized DenseNet121

The confusion matrix in Figure 7 shows that the Optimized DenseNet 121 model obtains outstanding class-wise classification scores in all categories of grape leaves. Precisely, Bacterial Rot was accurately detected in 14/15 samples, Downy Mildew in 142/146, Healthy Leaves in 187/188 and Powdery Mildew in 56/61 samples. There were few misclassifications, and they only took place between visually similar classes, like Powdery Mildew and Downy Mildew. Such findings underscore the ability of the model to adequately describe both slight and unique attributes of each group, and it shows the strength and consistency in the ability to classify leaves.

### D. Computational Cost Analysis of All evaluated models

Figure 8 shows the comparison of inference and training time of all the models compared. The Optimized DenseNet121 continues to perform better compared to the other models with the lowest inference and training costs. This efficiency combined with high accuracy, makes DenseNet121 especially adequate to real-world uses where performance and computational cost are most important. Despite the competitive performance of models such as SquezeeNet and AlexNet, the results of DenseNet121 provide a more suitable trade-off between speed and accuracy.

TABLE III
TESTING MODEL PERFORMANCE COMPARISON

| Model | Accuracy | Recall | Precision | F1 Score | MCC | PR AUC | Kappa | Specificity |
|---|---|---|---|---|---|---|---|---|
| ResNet18 | 97.56% | 97.81% | 97.81% | 97.80% | 96.57% | 99.56% | 96.56% | 99.21% |
| VGG16 | 96.34% | 96.34% | 96.34% | 96.34% | 94.93% | 98.63% | 91.21% | 97.90% |
| AlexNet | 94.39% | 94.39% | 94.35% | 94.36% | 91.21% | 98.63% | 91.21% | 97.90% |
| SqueezeNet | 95.12% | 95.12% | 95.12% | 95.12% | 94.03% | 98.62% | 90.76% | 97.73% |
| **Proposed Optimized DenseNet121** | **99.27%** | **99.27%** | **99.29%** | **99.28%** | **98.86%** | **99.99%** | **98.86%** | **99.71%** |

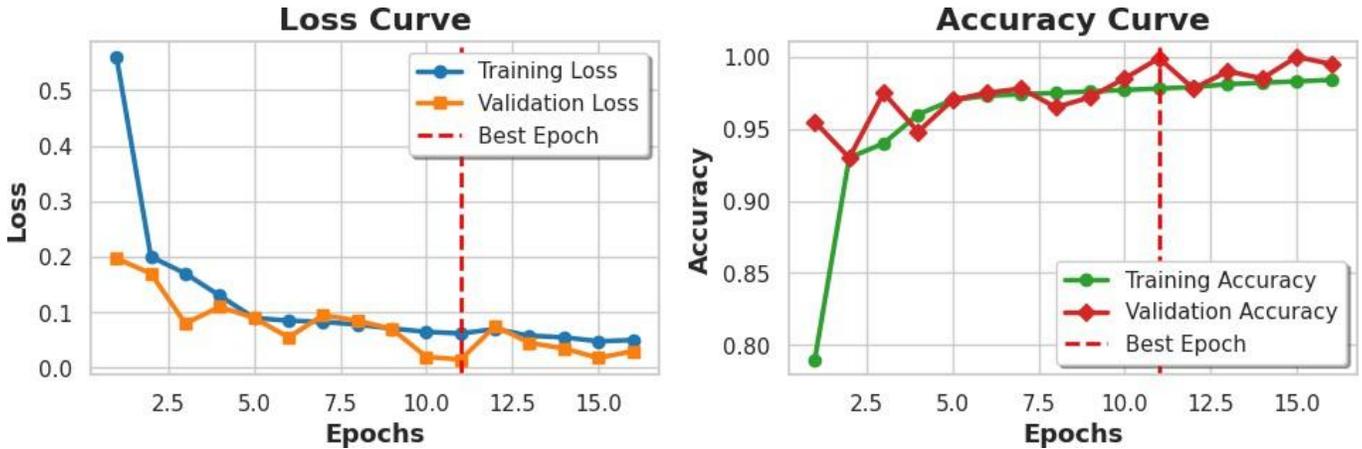

Fig. 6. Convergence behavior of the optimized DenseNet121 model where loss curves rapidly decrease before stabilizing.

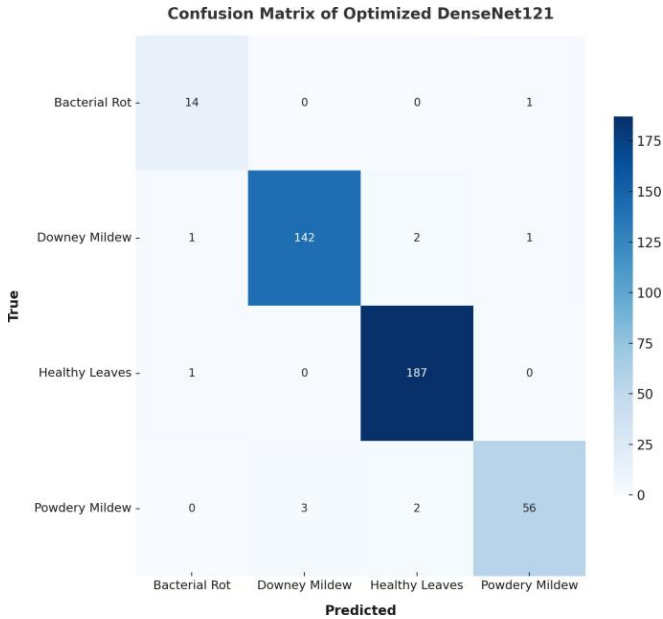

Fig. 7. The optimized DenseNet121 test set confusion matrix (high diagonal values indicating high accuracy and low off-diagonal entries indicating low misclassifications).

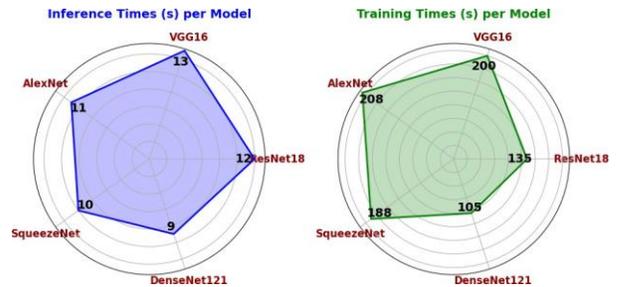

Fig. 8. Comparison of inference and training time for all evaluated models. Optimized DenseNet121 shows the lowest training and inference time.

### E. Proposed Optimized DenseNet-121 Interpretability Using Grad-CAM

To have a more insightful look at how the proposed Optimized DenseNet121 makes its predictions, we used Grad CAM to visualize the parts of the grape leaf that the model gave the highest importance. Figure 9 demonstrates that the emphasized regions represent disease specific characteristics of lesions, spots, and discolorations, which are important in differentiating between classes. The visualization will show that the model is rather effective to concentrate on the most informative areas instead of the background areas. The highlighted areas are verified by manual inspection and shown to be consistent with biologically relevant features, showing the interpretability, reliability, and strength of the decision-making process in the model.

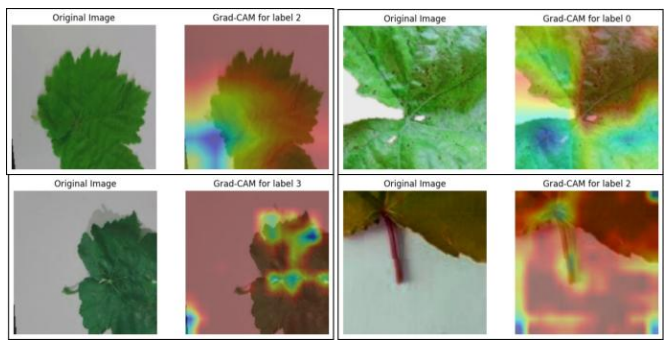

Fig. 9. Grad-CAM visualization of Optimized DenseNet-121

*F. Comparative Analysis with Existing Studies*

Table IV compares the proposed Optimized DenseNet121 with several representative models for grape leaf disease detection. Prior works have explored enhancing DenseNet architectures with attention mechanisms such as CBAM and SE [30], which improved feature extraction but were evaluated on relatively modest datasets. Similarly, Shetty et al. [25] demonstrated the advantages of DenseNet over standard CNNs, reporting 97.00% accuracy but with limited field-scale validation. Jin et al. [31] investigated CNN/DenseNet variants with hyperspectral imaging, achieving approximately 95.00% accuracy while highlighting the benefits of spectral-spatial feature extraction, though at the expense of costly equipment and higher computational demand. Lin et al. [32] introduced GrapeNet, an enhanced DenseNet121 with RFFB and CBAM modules, achieving 86.29% accuracy, which outperformed the DenseNet121 baseline (84.77%) while reducing parameters and training time.

In comparison, our proposed Optimized DenseNet121 achieves 99.27% accuracy, establishing a clear performance margin over prior DenseNet-based and hyperspectral approaches. The framework integrates domain-specific preprocessing and Grad-CAM interpretability, enabling biologically meaningful visualization of disease regions and enhancing model transparency. Architectural refinement ensures efficient computation, making the model suitable for real-time deployment in vineyard management systems.

Our proposed Optimized DenseNet121 sets a new benchmark for grape leaf disease detection by combining state-of-the-art accuracy, computational efficiency, and explainability. This positions the model as a practical and scalable solution for precision viticulture. We believe it has the ability to directly influence disease monitoring and management at both research and field levels.

## V. Conclusion

This paper introduces an Optimized DenseNet121 architecture of grape leaf disease detection, which overcomes the drawbacks of computational cost, generalisation, and interpretability of the current techniques. Domain specific preprocessing focuses on disease-relevant features, whereas dense connectivity focuses on the low-level structure, including veins, and high-level structure, including lesions. Extensive comparison with baseline CNNs proves better performance, at 99.27% accuracy, 99.28% F1 score, 99.71% specificity, 98.86% Kappa, and inference time of 9 seconds, with cross-validation revealing generalizability on different classes. Moreover, the addition of Grad-CAM offers visual explanations of the decisions made by the model, identifies biologically relevant regions and increases the confidence. Optimization minimizes computational expense and makes deployment possible in real time, whereas transfer learning guarantees consistent performance on small and unbalanced datasets. In general, the framework provides a scalable, precise, and practical solution, which can seal the research gaps and promote the application of practical solutions in various vineyard settings. Future research may consider the combination of multispectral imaging and real-time edge deployment to enhance the framework and make it even more robust and applicable to a variety of vineyard settings.

TABLE IV
COMPARISON OF EXISTING RECENT MODELS WITH THE PROPOSED OPTIMIZED DENSENET121 FOR GRAPE LEAF DISEASE DETECTION

| Study | Model/Architecture | Accuracy (%) | Pros | Cons/Limitations | Year |
|---|---|---|---|---|---|
| [30] | DenseNet121 + CBAM + SE attention modules | 96.40 | Attention-augmented DenseNet121 improves feature extraction; lightweight enhancement; strong performance on grape leaf diseases | Accuracy lower than some hybrid CNN/Transformer models; dataset modest in size; lacks vineyard-scale validation | 2025 |
| [25] | DenseNet (compared with CNN baseline) | 97.00 | Better feature extraction vs. plain CNN; demonstrates DL feasibility in viticulture; simple design | Evaluated on limited dataset; lacks field-scale validation; risk of overfitting; exploratory conference-level study | 2025 |
| [31] | CNN/DenseNet variants with hyperspectral feature extraction | ≈ 95.00 | Hyperspectral imaging enables early disease detection; field-collected data; DenseNet strong in spectral-spatial feature extraction | Requires expensive hyperspectral equipment; relatively small dataset; higher computational cost than RGB methods | 2022 |
| [32] | DenseNet121 baseline vs. GrapeNet (RFFB + CBAM) | 84.77 | DenseNet121 provided strongest baseline among CNNs; GrapeNet further improved accuracy (+1.52%), halved training time, and reduced parameters by 4.8M | DenseNet121 struggled with fine-grained disease stages; dataset relatively small (2850 images); no external vineyard-scale testing | 2022 |
| Proposed Study | Optimized DenseNet121 (with domain preprocessing + Grad-CAM interpretability) | 99.27 | High accuracy; efficient computation; explainability via Grad-CAM; suitable for real-time deployment | Needs extensive field-scale validation; tested mainly on curated datasets | 2025 |